\documentclass[conference]{IEEEtran}

\usepackage{amsmath,amssymb}
\usepackage{graphicx}
\usepackage{booktabs}
\usepackage{multirow}
\usepackage{xcolor}
\usepackage{url}
\usepackage{flushend}

\usepackage[
 backend=biber
,style=ieee
,minbibnames=1
,maxbibnames=2
,mincitenames=1
,maxcitenames=2
,eprint=true
,doi=true
,isbn=false
,url=true
]{biblatex}
\bibliography{references}
\AtBeginBibliography{\footnotesize}

\DeclareSourcemap{
  \maps{
    \map{
      \step[fieldset=month, null]
      \step[fieldset=address, null]
      \step[fieldset=location, null]
      \step[fieldset=publisher, null]
      \step[fieldset=url, null]
      \step[fieldset=isbn, null]
      \step[fieldset=editor, null]
    }
  }
}

\usepackage[
hidelinks,
  colorlinks=true,
  citecolor=blue,
  pdfstartview=Fit,
  breaklinks=true
]{hyperref}

\usepackage{cleveref}
\Crefname{figure}{Fig.}{Figs.}

\title{Feature Suppression and Differential Privacy for Residential Traffic Classification: A Two-Home Federated Study}
\author{
  \IEEEauthorblockN{M\'arton P\'al Lipcsey-Magyar\IEEEauthorrefmark{1} and
                    Adrian Pekar\IEEEauthorrefmark{1}\IEEEauthorrefmark{2}}
  \IEEEauthorblockA{\IEEEauthorrefmark{1}Budapest University of Technology and Economics, Hungary}
  \IEEEauthorblockA{\IEEEauthorrefmark{2}CUJO LLC Hungary}
  \IEEEauthorblockA{E-mail: lipcsey-magyarmartonpal@edu.bme.hu, apekar@hit.bme.hu}
}

\begin{document}

\maketitle

\begin{abstract}
Residential traffic classification supports service management, but learning across homes must account for heterogeneous traffic and privacy constraints. Privacy-aware training may impose uneven costs across traffic categories. We study this tradeoff in simulated two-client federated learning using 1.62 million preprocessed gateway-collected flows across six categories. We compare a full-feature baseline, feature suppression (FS), and differentially private stochastic gradient descent (DP-SGD) under one fixed record-level privacy setting. FS-mild excludes four timing features from 16 model inputs; it provides no formal privacy guarantee. With size-proportional aggregation, FS-mild achieves higher combined macro-F1 and worst-group F1 (the minimum per-class F1 across homes) than DP-SGD in all five seeds at both model capacities under stratified and temporal splits. The tested DP-SGD configuration incurs pronounced minority-category losses, especially in the smaller home, but FS-mild does not uniformly improve on the full-feature baseline. On stratified-split models, loss-based and shadow-model membership probes show near-chance aggregate discrimination without a consistent ranking across probes; this does not establish equivalent privacy. These findings support FS as an input-minimization baseline, not a substitute for formal privacy.
\end{abstract}


\section{Introduction}
\label{sec:intro}

Home gateways are useful vantage points for service and application management: traffic classification can support troubleshooting, parental controls, policy enforcement, and quality-of-service decisions. Building reliable classifiers requires diverse training examples, but centralizing household flow records raises privacy concerns. Federated learning (FL)~\cite{McMahan2017} enables clients to collaborate through model updates rather than uploading raw training records. Model exchange alone, however, does not guarantee privacy.

For service operators, the long tail matters. A classifier that recognizes dominant Web and Network traffic but fails on less frequent categories may miss precisely the traffic relevant to a troubleshooting task. This makes minority-category performance an operational concern beyond aggregate classification quality.

Households differ in application mix, volume, and minority-class coverage, producing statistical heterogeneity (non-IID data). Differential privacy (DP)~\cite{Dwork2013} provides a formal framework for bounding the influence of a protected unit on released outputs. Differentially private stochastic gradient descent (DP-SGD)~\cite{Abadi2016} implements private training through per-example gradient clipping and noise addition, but its utility cost can be uneven under heterogeneity~\cite{Bagdasaryan2019, Cheng2024RDPFL, Amiri2022FairnessDP, Du2025UnfairnessDP}. A distinct design choice is \emph{feature suppression} (FS): excluding selected flow features from model inputs. Our mild variant, FS-mild, removes four packet inter-arrival-time features from a 16-feature baseline. This minimizes the model's inputs, but does not by itself prevent collection of those features or provide a formal privacy guarantee.

We study these choices using traffic collected directly on two residential OpenWrt gateways. Each home's records form a client partition; FL training and aggregation are simulated on a common compute host, not executed on the gateways. We ask: \emph{for privacy-aware residential traffic analytics, what operational cost does DP-SGD impose relative to feature suppression in a small non-IID federation?} We evaluate one fixed DP-SGD privacy budget and clipping norm rather than a tuned privacy--utility frontier. Its protected unit is one training flow, not a user or household. The comparison evaluates classification utility and host-side training cost; it neither compares mechanisms at equivalent privacy protection nor measures gateway deployment performance.

Our contribution is a measurement and design-guidance study, not a new FL algorithm. Specifically:
\begin{enumerate}
    \item We characterize 1.62\,M preprocessed flows across six traffic categories from two homes, with different dominant categories and minority-category shares differing by up to 8.1$\times$. These captures provide a naturally heterogeneous setting, not a representative residential benchmark.
    \item We compare baseline FL, FS-mild, and DP-SGD across five seeds, two model capacities, and stratified and temporal splits. Evaluation covers macro-F1, worst-group F1 (the minimum per-class F1 across the two homes), and per-class behavior. Small-model secondary analyses cover stronger suppression (FS-aggressive), equal-weight aggregation, a FedDPA adaptation, and isolated runtime.
    \item With size-proportional aggregation, FS-mild exceeds the tested DP-SGD configuration on combined macro-F1 and worst-group F1 in every seed at both capacities under both splits. Class-level costs are uneven, with pronounced losses in the smaller home's minority categories. Loss-based and shadow-model membership inference on the stratified-split models yields near-chance aggregate AUC without a consistent configuration ranking across probes. These results indicate limited measured attack success, not evidence of equivalent privacy or of privacy gains from FS.
\end{enumerate}

The remainder of this paper is organized as follows. \Cref{sec:data} describes the two-home dataset and its non-IID structure; \Cref{sec:methods} presents the federated pipeline, privacy mechanisms, and evaluation protocol; \Cref{sec:results} reports the utility, worst-group, and membership-inference results; \Cref{sec:rw} surveys related work; and \Cref{sec:discussion,sec:conclusion} draw the lessons, limitations, and conclusions.

\section{Dataset and Setting}
\label{sec:data}

We evaluate bilateral FL using traffic captured at two residential home gateways. Each home's records form a separate client partition. Training and aggregation are simulated on a common compute host, not executed on the gateways; within the simulated protocol, aggregation uses model updates rather than raw flow records.

\subsection{Collection and Filtering}
The gateways run OpenWrt~\cite{OpenWrt}, a Linux-based operating system for embedded devices, and use a custom flow metering tool backed by the nDPI~5.0 classification library~\cite{nDPI2025}. The routers generated bidirectional flow records directly, with 80 fields including statistical features and application/category labels. The released Parquet files retain 78 fields after removing the source and destination IP addresses; they are not anonymous, as MAC addresses and timestamps remain. We use category labels rather than a fine-grained task with sparsely represented application labels. Recorded timestamps span 11.9 days for Home~A (Feb.~11--23, 2026) and 11.5 days for Home~B (Feb.~24--Mar.~8, 2026), using UTC dates.

We first require at least two bidirectional packets and retain records marked as deep packet inspection (DPI) classifications by the meter. These are inferred labels, not independently verified ground truth. We define the six-category task using a support rule: each category must have both ${>}0.4\%$ share of the packet- and DPI-filtered records and ${>}1{,}000$ flows in each home. The resulting fixed list is Network, Web, System, Media, Collaborative, and SocialNetwork; all other categories are excluded. This restricts the task without balancing the retained classes.

The final dataset contains 996\,450 flows from Home~A and 619\,284 from Home~B, retaining 84\% of the 1\,928\,110 captured records. The full-feature baseline represents each flow with 16 compact flow features, including both bidirectional statistics and directional counters, as described in \Cref{sec:methods}. The artifact records the counts at each filtering stage.

\begin{table}[t]
\centering
\caption{Capture spans and class distribution after filtering. Ratio = larger share / smaller share. FedAvg weights are fixed aggregation coefficients derived from training-set sizes, not learned model parameters.}
\label{tab:dataset}
\small
\setlength{\tabcolsep}{3.5pt}
\begin{tabular}{@{}lrrrrr@{}}
\toprule
Category & \multicolumn{2}{c}{Home A} & \multicolumn{2}{c}{Home B} & Ratio \\
\cmidrule(lr){2-3}\cmidrule(lr){4-5}
 & Count & \% & Count & \% & \\
\midrule
Network       & 547\,179 & 54.9 & 254\,471 & 41.1 & 1.3$\times$ \\
Web           & 326\,576 & 32.8 & 336\,632 & 54.4 & 1.7$\times$ \\
System        &  45\,533 &  4.6 &   4\,417 &  0.7 & 6.4$\times$ \\
Media         &  38\,800 &  3.9 &   2\,991 &  0.5 & 8.1$\times$ \\
Collaborative &  21\,258 &  2.1 &  14\,054 &  2.3 & 1.1$\times$ \\
SocialNetwork &  17\,104 &  1.7 &   6\,719 &  1.1 & 1.6$\times$ \\
\midrule
\textbf{Total} & \textbf{996\,450} & & \textbf{619\,284} & & \\
Capture span & \multicolumn{2}{c}{11.9 days} & \multicolumn{2}{c}{11.5 days} & -- \\
FedAvg weight & \multicolumn{2}{c}{61.7\%} & \multicolumn{2}{c}{38.3\%} & -- \\
\bottomrule
\end{tabular}
\end{table}

\subsection{Natural Non-IID Structure}
\Cref{tab:dataset} summarizes three forms of observed heterogeneity. First, the dominant categories differ: Network dominates Home~A, while Web dominates Home~B. Second, minority-category shares are highly skewed; System and Media differ by 6.4$\times$ and 8.1$\times$. Third, Home~A has 1.6$\times$ as many retained flows, giving it a 61.7\% size-proportional aggregation weight. Because the capture windows do not overlap, these differences combine household and time-window effects.

After filtering, Home~A has 21 distinct source MAC addresses and Home~B has 10; these are address counts, not verified device or user inventories. Home~A's Media records predominantly carry YouTube and RTSP labels, whereas Home~B's Media records are predominantly YouTube-labeled. Slack and GitHub labels occur in both homes. These observations characterize the captured traffic, not a controlled inventory of household devices or activities.

\section{Methods}
\label{sec:methods}

\subsection{Federated Pipeline}
We use FedAvg~\cite{McMahan2017} as an operational baseline with both homes participating in all 20 rounds. The three primary configurations---baseline FL, FS-mild, and DP-SGD---use 5 local epochs per round with a fresh Adam optimizer (lr$\,{=}\,10^{-3}$, nominal batch size 256). Aggregation is size-proportional by default; for the small model, we also repeat these three configurations with equal weights ($w_A{=}w_B{=}0.5$). FedDPA's two-stage local schedule is specified below. We do not claim FedAvg is optimal for non-IID data; non-IID-specific aggregation methods are outside the scope of this study.

The small multilayer perceptron (MLP) has hidden dimensions $[16,16]$; the medium robustness model uses $[128,64]$. Both use ReLU activations and cross-entropy loss. With all 16 inputs, they have 646 and 10\,822 trainable parameters, respectively. Hidden widths remain fixed within each capacity when features are suppressed; only the input layer loses weights. We retain the natural class imbalance without oversampling or class-weighted loss.

Each home is split independently. Main experiments use stratified 80/20 train--test partitions and five seeds. For both capacities, the three primary configurations also undergo an earlier-80\%/later-20\% split ordered by flow start time within each home. This tests temporal transfer but does not hold out devices or group related sessions. \Cref{tab:hyperparams} summarizes the training setup.

Preprocessing is record-local and uses no fitted statistics. Duration and packet inter-arrival time (PIAT) values are converted from milliseconds to seconds, and byte counts and packet-size statistics to KiB; these features and packet counts are transformed with $\log(1+x)$. The protocol number is divided by 255. The constants are fixed independently of both datasets, so altering one record does not change the representation of other records. All configurations use this same transformation before selecting their input columns.

\begin{table}[t]
\centering
\caption{Training setup. FedDPA uses five epochs in each of two local stages.}
\label{tab:hyperparams}
\small
\begin{tabular}{@{}ll@{}}
\toprule
Parameter & Value \\
\midrule
Small / medium MLP & $[16,16]$ / $[128,64]$ \\
Rounds / local epochs per stage & 20 / 5 \\
Batch size / optimizer & 256 / Adam, lr=$10^{-3}$ \\
Train--test split & 80/20, stratified or temporal \\
Seeds & 42, 123, 456, 789, 1024 \\
Target $\varepsilon$ / clip norm $C$ & 8.0 / 1.0 \\
$\delta$ (approx.), Home A / Home B & $(1.25,\ 2.02)\times10^{-7}$ \\
\bottomrule
\end{tabular}
\end{table}

\subsection{Privacy Mechanisms}
From the 80 captured fields, the baseline feature set uses 16 compact flow features suited to a lightweight model: duration, bidirectional and directional packet/byte counts, packet-size statistics, PIAT statistics, and protocol number.

\textbf{Feature suppression.} FS-mild removes the four PIAT features, reducing the input from 16 to 12 features. The choice is motivated by potential behavioral information in timing and the high zero rates of PIAT minima and standard deviations (37--61\% across homes). Neither motivation establishes that these features leak private behavior: we do not measure feature-specific leakage or compare against random feature removal. Suppression here excludes features from model inputs; avoiding their persistent collection would additionally require configuring the meter accordingly. FS-aggressive removes the four directional packet/byte counters as well, leaving 8 inputs, and is a secondary check on more extensive suppression.

\textbf{DP-SGD.} Differentially private stochastic gradient descent~\cite{Abadi2016} uses Opacus~\cite{Opacus2021} with per-example norm clipping, Gaussian noise, Poisson sampling, and R\'enyi DP accounting~\cite{Mironov2017}. The protected unit is one training flow under add/remove adjacency, not a device, user, or household. One accountant per home persists across all rounds. For local training size $N$, the actual sampling rate is $1/\lceil N/256\rceil$; calibration covers all $20\times5\times\lceil N/256\rceil$ steps. Noise targets $\varepsilon{=}8$ with $\delta{=}\min(10^{-5},1/(10N))$, yielding multipliers approximately 0.53 for Home~A and 0.56 for Home~B. We use one fixed privacy budget and clipping norm without DP-specific tuning or a privacy--utility sweep; measured costs therefore concern this configuration, not an intrinsic limit of DP.

\textbf{FedDPA adaptation.} The secondary small-model baseline follows Fisher-based personalization~\cite{Yang2023FedDPA}. Each round, the mean squared per-example gradients on up to 5{,}000 local examples estimate the diagonal Fisher. After per-tensor min--max normalization, values at least $\tau{=}0.4$ select personal coordinates retained from the previous local model; other coordinates start from the global model. Personal and shared coordinates are trained in separate five-epoch stages with fixed constraint coefficients $\lambda_1{=}\lambda_2{=}0.05$. Thus FedDPA performs ten local passes per round, not five.

In this adaptation, each client's entire update is clipped to $C{=}1$ and every coordinate is noised before aggregation. Independently observable client releases use replacement sensitivity $2C$ and per-coordinate noise standard deviation $2C\sigma$, calibrated over 20 full-participation releases. Accounting protects replacement of one client's training dataset, conditional on fixed roster and weights, rather than one flow. The same numeric $\varepsilon$ and per-home $\delta$ as DP-SGD do not imply equivalent privacy protection. FedDPA utility is evaluated on each home's local post-training model; other configurations use the final global model. The artifact specifies the adaptive constraints and release assumptions; this is a fixed-setting adaptation, not an optimally tuned reproduction.

\subsection{Evaluation Protocol}
We report macro-F1 and worst-group F1 at the final round. Macro-F1 is the unweighted average of per-class F1 within each home, then averaged equally across homes regardless of aggregation weights. Worst-group F1 is the minimum over all home/class pairs within each seed, followed by averaging across seeds. Paired 95\% intervals use seed-aligned differences, Student's $t$ distribution, and sample standard deviations. They quantify variability across these runs, not population uncertainty across households. Round time includes both homes' training, aggregation, and evaluation.

Two membership inference attacks (MIA) evaluate the exact final checkpoints of the three primary configurations under the stratified split, at both capacities. The loss-based probe uses negative cross-entropy as its membership score~\cite{Yeom2018}. The shadow probe trains four shadow FL models per target and fits per-home, per-class logistic regression over loss, confidence, entropy, true-class probability, and margin~\cite{Shokri2017}. Each shadow trains on a random half of each home's target-training partition; the remaining half supplies its nonmembers. Shadows use the corresponding features, architecture, and training mechanism, with DP noise recalibrated to their sample sizes. Target-test records are excluded from fitting shadows and attack classifiers. This is a controlled diagnostic with training-pool access, not an independently sourced auxiliary-data attack; the shadow/target size mismatch can affect transfer.

Both probes evaluate equal numbers of members and nonmembers per home/class, capped at 2{,}000 each, then macro-average across classes and homes. We report AUC, TPR@1\%FPR, TPR@5\%FPR, and maximum empirical $\mathrm{TPR}-\mathrm{FPR}$. ROC operating points are descriptive, not independently calibrated attack thresholds. Near-chance scores or intervals spanning zero do not establish privacy equivalence.

\subsection{Claim Scope and Reproducibility}
The probes infer membership from final-model outputs for labeled candidate flows; they do not audit an aggregation server observing intermediate updates. DP accounting concerns the training mechanisms conditional on fixed partitions and public protocol metadata, including roster, sizes, and aggregation weights. It does not cover data-dependent filtering or splitting, publication of evaluation metrics or datasets, or joint release of all runs; FedDPA's personalized models and metrics are also outside its update-release budget. We use seeded pseudorandomness for reproducible research simulations, not secret cryptographic randomness for a production private release. Reported $(\varepsilon,\delta)$ values characterize the accounted mechanisms under their randomness assumptions, not end-to-end privacy of the public artifact. FS has no formal privacy guarantee, and this is not an equal-protection comparison.

The artifact~\cite{Artifact} provides the pipeline, datasets, per-round logs, global target and shadow checkpoints, and analysis scripts. Dataset, partition, source, and checkpoint hashes link the measurements to their generating protocol; library versions and seeds are recorded. Deterministic execution is requested, but numerical identity across hardware and software environments is not assumed.

\section{Results}
\label{sec:results}

\begin{table}[t]
\centering
\caption{Small-model utility (mean $\pm$ sample std, 5 seeds). Combined macro-F1 averages homes equally; worst-group takes the minimum over home/class pairs within each seed. FS-mild/FS-aggr.\ retain 12/8 features (12f/8f). FedDPA uses personalized evaluation and a different privacy unit; other rows evaluate the global model.}
\label{tab:main_results}
\small
\setlength{\tabcolsep}{3pt}
\begin{tabular}{@{}lcccc@{}}
\toprule
Config & A Macro & B Macro & Combined & Worst \\
\midrule
Baseline FL       & .671\tiny$\pm$.083 & .520\tiny$\pm$.016 & .596\tiny$\pm$.041 & .077\tiny$\pm$.051 \\
FS-mild (12f)     & .714\tiny$\pm$.037 & .524\tiny$\pm$.013 & .619\tiny$\pm$.017 & .071\tiny$\pm$.044 \\
FS-aggr.\ (8f)    & .683\tiny$\pm$.043 & .524\tiny$\pm$.011 & .603\tiny$\pm$.026 & .041\tiny$\pm$.049 \\
DP-SGD            & .616\tiny$\pm$.039 & .389\tiny$\pm$.033 & .503\tiny$\pm$.034 & .004\tiny$\pm$.002 \\
FedDPA            & .509\tiny$\pm$.026 & .324\tiny$\pm$.020 & .416\tiny$\pm$.022 & .009\tiny$\pm$.015 \\
\bottomrule
\end{tabular}
\end{table}

\subsection{Utility and Worst-Group Performance}
\Cref{tab:main_results} gives the small-model comparison. FS-mild achieves higher macro-F1 than DP-SGD in both homes and higher worst-group F1. Relative to baseline FL, FS-mild has higher mean combined macro-F1 but slightly lower mean worst-group F1: suppressing timing features is not a uniform improvement over using all features. Baseline FL shows substantial between-seed variation on Home~A; FS-mild's higher combined mean does not establish a consistent advantage over the full-feature baseline. DP-SGD's macro-F1 reduction relative to the baseline is larger in Home~B than in Home~A.

FS-mild exceeds DP-SGD on both combined macro-F1 and worst-group F1 in every seed. The paired intervals exclude zero for both metrics (\Cref{tab:paired_evidence}). This supports an advantage over the fixed DP-SGD configuration tested here, not superiority over tuned DP or an equal-privacy alternative.

\subsection{Distribution of the Utility Cost}
\Cref{tab:perclass} shows that the utility cost is not uniform. Network and Web retain high F1 under DP-SGD, while Home~B's Media and SocialNetwork scores approach zero and its System score is roughly half the baseline's. Minority-category degradation is not confined to Home~B: Home~A also loses Collaborative and SocialNetwork performance. Conversely, Home~A Media improves over baseline FL under DP-SGD. These differences caution against treating noise as uniformly harmful to every class.

\begin{table}[t]
\centering
\caption{Per-class F1 (mean, 5 seeds) for the three primary small-model configurations (BL = Baseline FL, FS = FS-mild, DP = DP-SGD). Bold: highest mean within each home/class. Underline: below 0.10.}
\label{tab:perclass}
\small
\setlength{\tabcolsep}{2.5pt}
\begin{tabular}{@{}l ccc ccc@{}}
\toprule
 & \multicolumn{3}{c}{Home A} & \multicolumn{3}{c}{Home B} \\
\cmidrule(lr){2-4}\cmidrule(lr){5-7}
Class & BL & FS & DP & BL & FS & DP \\
\midrule
Collaborative & \textbf{.333} & .307 & \underline{.098} & .132 & \textbf{.153} & \underline{.074} \\
Media         & .466 & \textbf{.697} & .626 & \textbf{.163} & .108 & \underline{.021} \\
Network       & .975 & \textbf{.977} & .968 & .975 & \textbf{.977} & .963 \\
SocialNetwork & .453 & \textbf{.463} & .234 & \textbf{.247} & .233 & \underline{.007} \\
System        & .911 & \textbf{.940} & .884 & .666 & \textbf{.725} & .332 \\
Web           & .892 & \textbf{.900} & .889 & .937 & \textbf{.946} & .935 \\
\bottomrule
\end{tabular}
\end{table}

\begin{figure}[t]
\centering
\includegraphics[width=\columnwidth]{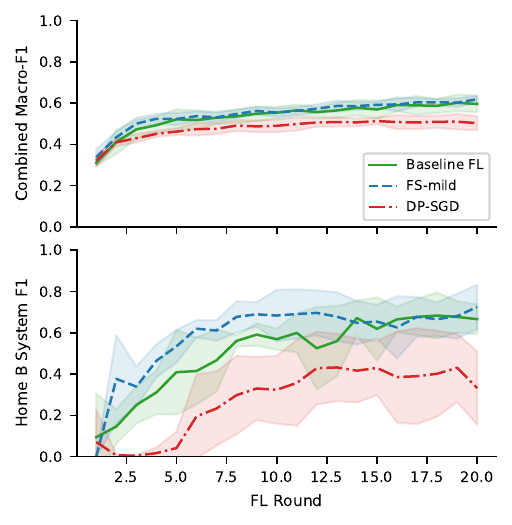}
\caption{Small-model convergence under the stratified split (mean $\pm$ sample std, 5 seeds). Top: combined macro-F1. Bottom: Home~B System F1. DP-SGD reaches lower final scores; the curves alone do not identify the cause of the gap.}
\label{fig:convergence}
\end{figure}

The convergence curves in \Cref{fig:convergence} show lower combined macro-F1 under DP-SGD after the initial rounds. Its mean Home~B System F1 rises through the middle rounds but finishes below its peak, with substantial variation across seeds; baseline FL and FS-mild reach higher final scores. These trajectories describe the observed training behavior, but do not isolate clipping from noise or optimization effects.

FS-mild also has limits: its Home~B Media and SocialNetwork means are below the baseline's. FS-aggressive further lowers mean combined macro-F1 and worst-group F1 relative to FS-mild, although some individual classes improve. More extensive suppression therefore does not yield uniformly better classification.

\subsection{Robustness and Secondary Configurations}
With the medium model, mean combined macro-F1 and worst-group F1 improve for all three primary configurations. FS-mild remains close to baseline FL on both metrics and exceeds DP-SGD: combined macro-F1 is .746 versus .555, and worst-group F1 is .348 versus .070. The paired differences remain positive across all seeds (\Cref{tab:paired_evidence}). This supports the result beyond the smallest architecture, although gateway execution costs have not been measured.

The earlier-80\%/later-20\% split by flow start time within each home preserves the FS-mild--DP-SGD ordering for both capacities: both metrics favor FS-mild in all five seeds. The temporal check therefore supports transfer to a later window within these captures, not to unseen homes or devices. Under equal-weight aggregation, the small-model macro-F1 advantage also persists. Worst-group evidence is weaker in that ablation: four seeds favor FS-mild, but the paired interval includes zero.

\begin{table}[t]
\centering
\caption{FS-mild minus DP-SGD: mean difference [paired 95\% CI], 5 seeds. Main = stratified, size-proportional aggregation. Both metrics favor FS-mild in all seeds except equal-weight worst-group (4/5).}
\label{tab:paired_evidence}
\small
\setlength{\tabcolsep}{3pt}
\begin{tabular}{@{}lcc@{}}
\toprule
Setting & Combined macro-F1 & Worst-group F1 \\
\midrule
Small, main & .116 [.055, .178] & .067 [.015, .118] \\
Medium, main & .191 [.135, .247] & .279 [.214, .344] \\
Small, temporal & .093 [.065, .120] & .079 [.004, .154] \\
Medium, temporal & .175 [.162, .189] & .210 [.159, .262] \\
Small, equal-weight & .060 [.034, .087] & .014 [$-$.033, .061] \\
\bottomrule
\end{tabular}
\end{table}

The FedDPA adaptation learns a nontrivial classifier, but its combined macro-F1 is below the three primary small-model configurations and its worst-group F1 remains low (\Cref{tab:main_results}). Its client-dataset replacement guarantee, personalized evaluation, and two-stage training differ from DP-SGD. This fixed-setting result neither ranks the mechanisms at equivalent privacy nor demonstrates a fundamental limit of adaptive DP in bilateral FL.

\subsection{Runtime}
\Cref{tab:runtime} separates timing from the utility experiment: configurations are measured sequentially on an AMD EPYC 7702 CPU host with one training thread, using seed 42. Each round includes both clients' training, aggregation, and evaluation. Estimated run time is 20 times the median round time, not measured end-to-end latency. The five-seed utility and attack logs were collected under a ten-worker schedule and are not used for this timing table. FedDPA executes two five-epoch stages plus Fisher estimation per round, so its workload is not identical to the other configurations. These are host-side simulation costs, not gateway benchmarks.

Baseline FL and both FS variants have similar per-round costs. DP-SGD takes approximately $3.4\times$ as long per round as FS-mild in this isolated measurement. FedDPA also takes longer than the baseline, but its additional training stage and Fisher estimation prevent interpreting the difference as privacy overhead alone.

\begin{table}[t]
\centering
\caption{Isolated small-model runtime: median of 20 rounds, seed 42, one CPU training thread. Estimated run (min) $=$ median round time (s) $\times$ 20 / 60. FedDPA uses two local stages plus Fisher estimation.}
\label{tab:runtime}
\small
\setlength{\tabcolsep}{3pt}
\begin{tabular}{@{}lcc@{}}
\toprule
Config & Round time (s) & Est.\ run (min) \\
\midrule
Baseline FL   & 54.6 & 18.2 \\
FS-mild       & 54.7 & 18.2 \\
FS-aggressive & 54.7 & 18.2 \\
DP-SGD        & 184.7 & 61.6 \\
FedDPA        & 124.4 & 41.5 \\
\bottomrule
\end{tabular}
\end{table}

\subsection{Membership Inference Findings}
\Cref{tab:mia} reports loss-based and shadow-model MIA results for the small model. Aggregate AUC remains near chance for the three primary configurations at both capacities, with means between .500 and .503. All paired AUC intervals comparing FS-mild with baseline FL or DP-SGD include zero. Some unadjusted low-FPR intervals exclude zero, but do not establish a consistent ranking: in the small model, loss-based TPR@5\%FPR favors DP-SGD over FS-mild, whereas the shadow probe favors FS-mild. Full per-home and medium-model outputs are provided in the artifact.

These results establish limited measured attack success, not absence of leakage or equivalence of protection. The shadow probe uses access to the target-training pool, and neither probe tests intermediate-update exposure or temporal-split models. DP-SGD's accounted record-level guarantee is a distinct property even when these empirical probes offer little separation; FS has no such guarantee. The operational comparison is therefore a measured utility cost at one DP setting, not evidence that privacy protection has no benefit.

\begin{table}[t]
\centering
\caption{MIA metrics for the small model (equal-home average, mean $\pm$ sample std, 5 seeds). Loss = loss-based attack; Shadow = shadow-model attack. TPR columns use 1\%/5\% FPR; Adv.\ is maximum empirical TPR$-$FPR. Values are rounded to three decimals.}
\label{tab:mia}
\small
\setlength{\tabcolsep}{2.5pt}
\begin{tabular}{@{}llcccc@{}}
\toprule
Attack & Config & AUC & TPR@1\% & TPR@5\% & Adv. \\
\midrule
\multirow{3}{*}{Loss}
& Baseline FL & .501\tiny$\pm$.003 & .007\tiny$\pm$.001 & .042\tiny$\pm$.003 & .023\tiny$\pm$.005 \\
& FS-mild     & .502\tiny$\pm$.003 & .008\tiny$\pm$.000 & .039\tiny$\pm$.003 & .022\tiny$\pm$.005 \\
& DP-SGD      & .500\tiny$\pm$.002 & .006\tiny$\pm$.001 & .035\tiny$\pm$.004 & .021\tiny$\pm$.003 \\
\midrule
\multirow{3}{*}{Shadow}
& Baseline FL & .502\tiny$\pm$.001 & .010\tiny$\pm$.001 & .049\tiny$\pm$.003 & .023\tiny$\pm$.002 \\
& FS-mild     & .500\tiny$\pm$.003 & .008\tiny$\pm$.001 & .044\tiny$\pm$.003 & .021\tiny$\pm$.002 \\
& DP-SGD      & .501\tiny$\pm$.004 & .010\tiny$\pm$.001 & .049\tiny$\pm$.001 & .023\tiny$\pm$.003 \\
\bottomrule
\end{tabular}
\end{table}

\section{Related Work}
\label{sec:rw}

\begin{table*}[t]
\centering
\caption{Setting-specific operational guidance, not a ranking at equivalent privacy protection.}
\label{tab:guidance}
\small
\setlength{\tabcolsep}{3pt}
\begin{tabular}{@{}p{4.2cm}p{3.0cm}p{10.5cm}@{}}
\toprule
Deployment priority & Candidate & Evidence and boundary \\
\midrule
Training-flow privacy guarantee & DP-SGD & Record-level accounting under the assumptions in \Cref{sec:methods}; not end-to-end privacy of the artifact \\
Minority-category utility & FS-mild & Higher worst-group F1 than tested DP-SGD under size-proportional aggregation; not uniformly better than baseline FL \\
Low training-time overhead & FS-mild & Baseline-like isolated small-model runtime on the CPU host; no gateway timing evidence \\
Fewest retained inputs & FS-aggressive & Eight features; lower mean combined macro-F1 and worst-group F1 than FS-mild in the small model, without demonstrated privacy gain \\
Personalized DP & FedDPA adaptation & Secondary fixed-setting evidence with low worst-group F1; different privacy unit and evaluation from DP-SGD \\
\bottomrule
\end{tabular}
\end{table*}

\textbf{DP in heterogeneous FL.}
\citeauthor{Cheng2024RDPFL}~\cite{Cheng2024RDPFL} analyze how statistical heterogeneity affects the utility loss from clipping and noise, while \citeauthor{Xiong2022PrivacyThreat}~\cite{Xiong2022PrivacyThreat} study inference risk and propose local- and server-side noise addition for non-IID FL. Mitigation approaches include noise-aware aggregation in Robust-HDP~\cite{Malekmohammadi2024ICML}, Fisher-based personalization and adaptive constraints in FedDPA~\cite{Yang2023FedDPA}, and adaptive server optimization in AdDPNFL~\cite{Chen2024DPNFL}. These approaches show why one fixed DP-SGD configuration cannot characterize the best achievable DP utility. Our FedDPA adaptation is a secondary evaluation, not a comparison covering all these alternatives.

\textbf{Unequal utility costs.}
\citeauthor{Bagdasaryan2019}~\cite{Bagdasaryan2019} demonstrate disproportionate DP-SGD accuracy losses for underrepresented classes and subgroups. In FL, \citeauthor{Amiri2022FairnessDP}~\cite{Amiri2022FairnessDP} examine performance and fairness under non-IID data, while \citeauthor{Du2025UnfairnessDP}~\cite{Du2025UnfairnessDP} study and mitigate disparities in model performance across clients. Our home/category F1 analysis is an operational robustness measure, not a demographic fairness evaluation. It adds evidence from residential flow records about how the tested configurations distribute utility across traffic categories.

\textbf{Federated traffic classification.}
PrivPkt~\cite{Akbari2019PrivPkt} studies DP-SGD and shadow-model membership inference for packet-based encrypted traffic classification, including federated training. It reports greater membership vulnerability for smaller classes in its non-private, unbalanced setting and evaluates differential privacy and training-data balancing as mitigations. FedPacket~\cite{Bakopoulou2022FedPacket} uses HTTP keys rather than sensitive values in its federated packet-classification features. It also demonstrates feature inference by an honest-but-curious server, illustrating that feature restriction alone does not prevent leakage. FEAT~\cite{Guo2023FEAT} estimates traffic-data skewness and selects clients to improve classification under heterogeneity. Our study instead examines fixed feature suppression alongside DP-SGD on gateway-collected residential flow statistics in a simulated federation, emphasizing home/category classification utility and temporal transfer. It is a different task and protocol, not an accuracy ranking against these systems.

\textbf{Non-IID FL optimization.}
FedProx~\cite{FedProx2020} addresses statistical and systems heterogeneity, while SCAFFOLD~\cite{SCAFFOLD2020} uses control variates to correct client drift. We keep FedAvg fixed as an operational reference, rather than claim it is the best non-IID optimizer. Whether these alternatives narrow the measured FS-mild--DP-SGD gap requires a separate comparison.

\textbf{Feature selection and data minimization.}
\citeauthor{Hu2023PSO_KBS}~\cite{Hu2023PSO_KBS} use particle swarm optimization for federated feature selection with a trusted third party. FedSDG-FS~\cite{Li2024FedSDGFS} combines feature selection with local-embedding perturbation for DP in vertical FL. FedFed~\cite{Yang2023FedFed} distills and shares noised features to mitigate heterogeneity, evaluating membership inference as well as utility; it does not simply discard input columns. These works optimize feature selection or representation sharing. We instead evaluate fixed feature-family exclusions in horizontal FL, alongside DP-SGD and two membership probes. The distinction is the residential measurement question, not a new feature-selection algorithm or a demonstrated privacy guarantee from suppression.

\section{Discussion and Lessons}
\label{sec:discussion}

The answer to our research question is conditional on the tested training configuration. With size-proportional aggregation, DP-SGD yields lower combined macro-F1 and worst-group F1 than FS-mild at both capacities under both split protocols. Its accounted record-level protection is a distinct benefit, not something the near-chance MIA results invalidate. Runtime is assessed separately in \Cref{tab:runtime}; the utility differences should not be read as an equal-protection comparison. We draw three lessons for privacy-aware residential traffic analytics.

\textbf{Lesson 1: DP cost is class-dependent.}
DP-SGD does not lower all scores uniformly. In the primary small-model comparison, Home~B loses more macro-F1 than Home~A relative to baseline FL, with pronounced losses in minority categories while Network and Web remain strong. Some categories nevertheless improve relative to baseline FL. The slightly higher noise multiplier for Home~B does not by itself explain this pattern: clipping, sampling, optimization, and class composition are not separately varied. For service management, the practical lesson is to inspect per-category performance rather than choose a configuration solely by its aggregate score.

\textbf{Lesson 2: feature suppression is a practical baseline, not formal privacy.}
FS-mild uses fewer input features and has runtime close to baseline FL in the isolated small-model measurement. It performs better than the tested DP-SGD configuration, but does not uniformly improve on the full-feature baseline. Our experiments exclude timing features from learning; they neither demonstrate reduced feature-specific leakage nor remove those fields from the captured dataset. Avoiding their collection would require a corresponding metering policy. If formal protection is required, feature suppression alone is insufficient: the protected unit and all observable releases must be specified, and DP settings should be tuned against the application's utility requirements.

\textbf{Lesson 3: the tested MIA probes provide limited discrimination.}
Aggregate AUC remains near chance across the primary configurations and both capacities. The low-FPR comparisons sometimes distinguish configurations, but their directions are not consistent across probes. This supports limited measured attack success, not privacy equivalence or absence of memorization. The shadow probe's target-training-pool access and smaller shadow training sets constrain its interpretation; neither probe evaluates an observer of intermediate updates. Attack evaluation complements, rather than replaces, mechanism-level accounting.

\Cref{tab:guidance} condenses these lessons into deployment-oriented guidance.

\textbf{Limitations.}
The evidence comes from two homes, full participation, six retained categories, and two MLP capacities. Non-overlapping capture windows confound household and temporal effects; DPI-derived labels are not independently verified ground truth. Stratified flow splits may share related traffic across training and test sets. Both capacities retain the FS-mild--DP-SGD ordering under temporal splitting, but this tests only later windows within the same captures, not unseen devices, grouped sessions, or new households. Five-seed intervals quantify run variability, not population-level generality. Equal-weight aggregation retains the macro-F1 advantage, while its worst-group interval includes zero.

We use one DP-SGD budget and clipping norm without DP-specific tuning, and do not evaluate non-IID-specific aggregators or random-feature-removal controls. These choices leave open how much of the utility gap is avoidable and whether PIAT suppression is preferable to other feature choices. FedDPA is a fixed-setting adaptation with a different privacy unit and personalized evaluation, not an equal-protection comparison. Training is simulated on a compute host, and isolated timing uses one seed rather than a multi-host benchmark. The public datasets and seeded checkpoints are reproducibility artifacts, not an end-to-end DP release. These boundaries make the study evidence for this operating setting, not a universal recommendation against DP.

\section{Conclusion}
\label{sec:conclusion}

This study contributes a gateway-collected two-home dataset and a reproducible comparison of feature suppression and fixed-setting DP-SGD, with explicit boundaries on privacy protection and generalization. With size-proportional aggregation, FS-mild achieved higher combined macro-F1 and worst-group F1 than DP-SGD in every seed at both model capacities under stratified and temporal splits. The measured utility advantage does not establish equivalent privacy: the two MIA probes show limited aggregate discrimination, while FS provides no formal guarantee. Feature suppression is therefore a useful input-minimization baseline to evaluate, not a replacement for DP when formal protection is required. Next steps are DP-specific tuning, matched feature-removal controls, independent attack data, and validation across additional homes and non-IID-specific aggregators.

\section*{Acknowledgement}

We thank Bal\'azs Pej\'o for early discussions that motivated the comparison between feature suppression and differential privacy. We thank Gergely Bicz\'ok for comments on an earlier draft.

Supported by the CELTIC-NEXT project Robust and AI Native 6G for Green Networks (RAI6-Green, C2023/1-9), funded by the National Research, Development and Innovation Fund of Hungary under Grant-2024-1.2.6-EUREKA-2024-00009.

\printbibliography

\end{document}